\documentclass{article}
\usepackage{ijcai21}

\usepackage{times}

\usepackage{soul}
\usepackage{url}
\usepackage[hidelinks]{hyperref}
\usepackage[utf8]{inputenc}
\usepackage[small]{caption}
\usepackage{graphicx}
\usepackage{amssymb}
\usepackage{amsmath}
\usepackage{amsthm}
\usepackage{dsfont}
\usepackage{stmaryrd}
\usepackage{booktabs}
\usepackage{algorithm}
\usepackage{algorithmic}
\usepackage[switch]{lineno}
\usepackage{multirow}
\usepackage{subfigure}
\DeclareMathOperator*{\argmax}{argmax}

\newtheorem{rem}{Remark}
\newtheorem{prop}{Proposition}

\title{Two-stage Training for Learning from Label Proportions}

\author{
Jiabin Liu$^{1,4}$\footnote{Equal Contribution}
\and
Bo Wang$^{2}$\footnotemark[1]\footnote{Corresponding Authors}\and
Xin Shen$^{3}$\and
Zhiquan Qi$^4$\footnotemark[2]\And
Yingjie Tian$^4$
\affiliations
$^1$AI Lab, Samsung Research China - Beijing\\
$^2$University of International Business and Economics\\
$^3$The Chinese University of Hong Kong\\
$^4$University of Chinese Academy of Sciences
\emails
liujiabin008@126.com,
wangbo@uibe.edu.cn,
xshen@se.cuhk.edu.hk,\\
qizhiquan@foxmail.com,
tyj@ucas.ac.cn
}

\begin{document}

\maketitle

\begin{abstract}
Learning from label proportions (LLP) aims at learning an instance-level classifier with label proportions in grouped training data.
Existing deep learning based LLP methods utilize end-to-end pipelines to obtain the proportional loss with Kullback-Leibler divergence between the bag-level prior and posterior class distributions.
However, the unconstrained optimization on this objective can hardly reach a solution in accordance with the given proportions.
Besides, concerning the probabilistic classifier, this strategy unavoidably results in high-entropy conditional class distributions at the instance level.
These issues further degrade the performance of the instance-level classification.
In this paper, we regard these problems as noisy pseudo labeling, and instead impose the strict proportion consistency on the classifier with a constrained optimization as a continuous training stage for existing LLP classifiers.
In addition, we introduce the mixup strategy and symmetric cross-entropy to further reduce the label noise.
Our framework is model-agnostic, and demonstrates compelling performance improvement in extensive experiments, when incorporated into other deep LLP models as a post-hoc phase.
\end{abstract}

\section{Introduction}

Learning from label proportions (LLP) is an important weakly supervised classification problem with only the label proportions in grouped data available.
Still, training LLP aims to obtain an instance-level classifier for the new-come inputs.
Successfully resolving LLP problems greatly contribute to many real-life applications: demographic classification \cite{ardehaly2017co}, US presidential election \cite{sun2017probabilistic}, embryo implantation prediction \cite{hernandez2018fitting}, spam filtering \cite{kuck2012learning}, video event detection \cite{lai2014video}, visual attribute modeling \cite{chen2014object,yu2014modeling}, and traffic flow prediction \cite{liebig2015distributed}.

On the one hand, the learnability of LLP strongly depends on the instances grouping and the proportions distribution in the bags.
\cite{yu2014learning} studied the instance-level empirical proportions risk minimization (EPRM) algorithm for LLP, w.r.t. the number, the size, and the prior label distribution of the bags.
They proved that LLP is learnable with the EPRM principle and given the bound of expected learning risk.

On the other hand, EPRM strives to minimize bag-level label proportions error.
Normally, this goal is achieved by minimizing Kullback-Leibler (KL) divergence between prior and posterior class distributions in each bag.
However, bag-level proportional information hardly provides sufficient constraints to perfectly solve LLP, because too many instance-level classifiers can satisfy proportional constraints exactly.
In other words, when considering instance-level classification, LLP is ill-posed.
As a consequence of the underdetermination, despite a number of achievements have been developed to resolve LLP accurately, it is still of great importance to design effective instance-level learning scheme to significantly improve the performance on high-dimensional data, e.g., images, merely with the proportional information.


In this paper, we challenge this issue through explicit \emph{weakly supervised clustering} with proportions, instead of the inadequate bag-level KL divergence.
To be concrete, unsupervised clustering is expected to discover data clusters specifically corresponding to classes.
However, na\"ively applying clustering without supervision will result in a trivial solution to assign all the instances to one cluster, deviating from the semantics, i.e., classes.
Fortunately, we can avoid this degeneration by imposing appropriate constraints on the cluster distribution \cite{caron2018deep}.
For example, when there is no knowledge on labels, we impose discrete uniform distribution to labels, leading to equal clustering.
Besides, in a semi-supervised learning protocol, the clustering result can be restricted with the help of labeled instances \cite{asano2019self} as well.
Similarly, in LLP scenario, we draw inspiration from this constrained clustering, to leverage label proportions to constrain cluster sizes.
Specifically, we build a constrained optimization problem, regarding feasible solutions as pseudo-labels that accurately complying proportions.

\begin{figure}
\centering
\includegraphics[width=0.46\textwidth]{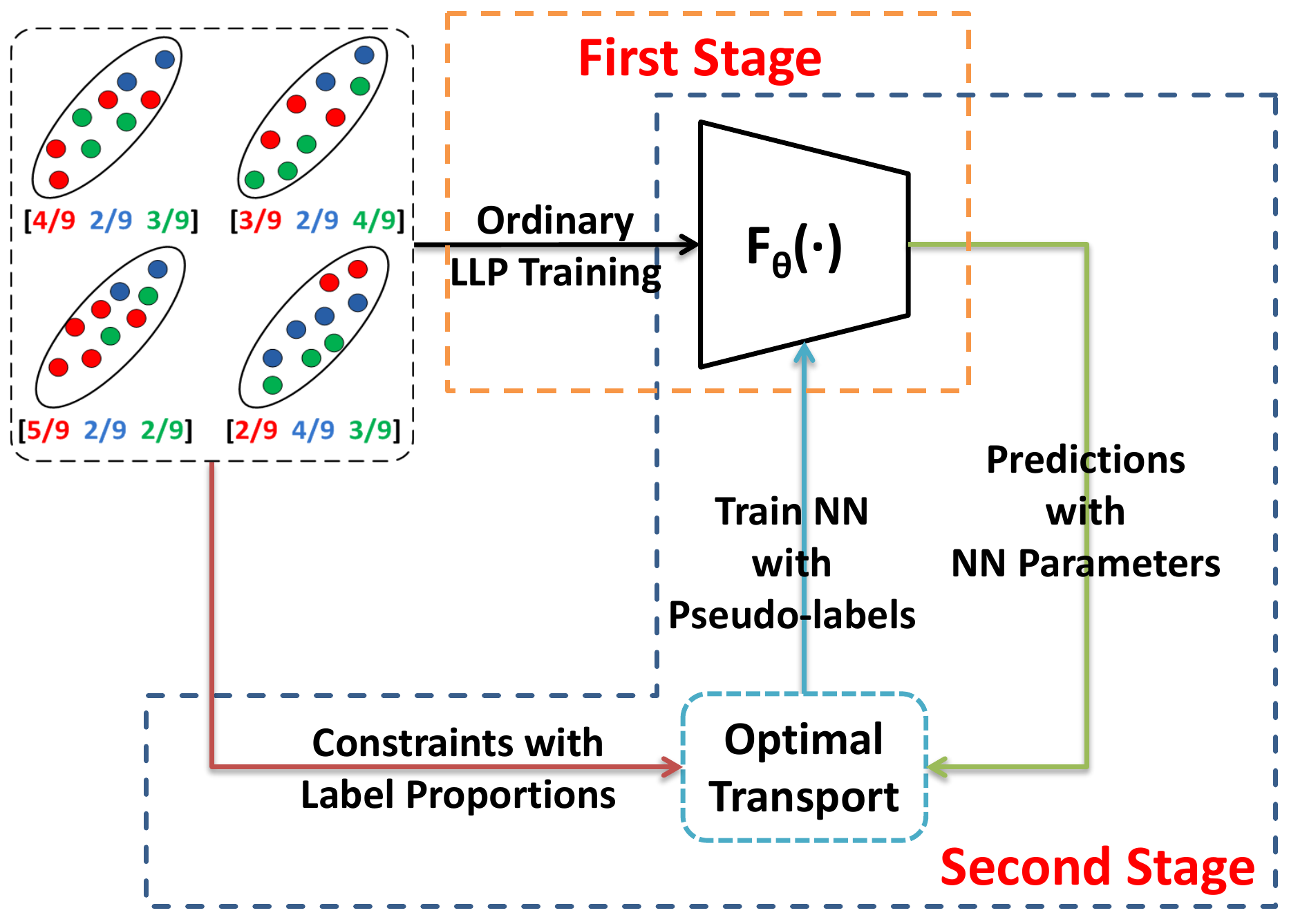}
\vspace{-5pt}
\caption{Illustration of the proposed two-stage LLP training framework. (Best viewed in color)}
\label{framework}
\vspace{-5pt}
\end{figure}
In order to properly tackle the weak supervision, we cast our framework, called PLOT, by considering pseudo labeling and the classification within one objective, utilizing label proportions as the constraints.
To implement the proposed schema, we alternately update the network parameters and the pseudo-labels, where optimal transport (OT) \cite{peyre2019computational} is employed to conduct the pseudo labeling process and standard cross-entropy minimization is adopted to train the network parameters.
Besides, to recognize the resulting pseudo-label as noisy label to the unknown clean one, we spur a new line to propose a \emph{two-stage} training strategy: Set our method as a following stage after other LLP solvers, and train the classification network with \emph{symmetric cross-entropy (SCE)} \cite{wang2019symmetric} along with \emph{mixup} dataset \cite{zhang2018mixup} to combat the memorization of noisy labels and inject the robustness into the solution.
We illustrate the proposed two-stage framework in Figure \ref{framework}.

In summary, our main contributions are four-fold:
\vspace{-3pt}
\begin{itemize}
\item We propose a novel LLP framework PLOT, which aims at meeting the exact proportions with a constrained optimization, and is an essentially orthogonal LLP treatment to the existing LLP algorithms;
\vspace{-3pt}
\item We conduct an alternate optimization process to solve the optimization on both the neural network classifier and pseudo labeling, by applying SCE to learn from the noisy pseudo-labels obtained with OT;
\vspace{-3pt}
\item Our framework is model-agnostic, and we demonstrate that it can easily fit for various deep-based LLP models as a post-hoc stage to further boost their performance;
\vspace{-3pt}
\item With lightly additional hyper-parameter and mixup data augmentation, our framework achieves state-of-the-art LLP performance on several benchmark datasets, based on the neural network pipeline with standard settings.
\end{itemize}

\section{Related Work}

To our knowledge, four end-to-end pipelines have been proposed for LLP, with deep neural networks as the backbone architecture.
Specifically, DLLP \cite{ardehaly2017co} is the first end-to-end LLP algorithm, with the KL divergence of prior and posterior proportions as the objective.
Although DLLP can learn a competent instance-level classifier, it is hardly in accordance with the proportions in training data, especially with large bag sizes (e.g., $>\!64$).
In order to ``guess'' instance labels that are consistent to  proportions \cite{felix2013psvm}, \cite{dulac2019deep} perform convex relaxation to labeling variables and \emph{first} recognize it as an OT problem.
Aiming at end-to-end training, they instead impose inexact labeling with KL divergence similar to DLLP, and efficiently solve the resulting unbalanced OT problem with entropic regularization.
However, the KL divergence results in inaccurate labeling, thus introduces additional noise to pseudo-labels compared to accurate labeling.
We discuss the differences between their work and our work in Section \ref{dc}.

Recently, LLP-GAN \cite{liu2019learning} greatly improves DLLP with adversarial mechanism.
In detail, the discriminator is designed as a ($K\!+\!1$)-way classifier, where the first $K$ classes indicate the real ones, and the $(K\!+\!1)^{th}$ class accounts for the generated one.
The main insight is to produce better representation with adversarial learning, thus boosting the downstream discriminative tasks.
In spite of substantial performance improvement compared with previous methods, LLP-GAN suffers from training instability, which inherits the characteristics from GANs \cite{gui2020review}.
Besides, in order to obtain satisfactory results, it requires subtle network design and hyper-parameters selection.
Similarly, based on visual adversarial training \cite{miyato2018virtual} in semi-supervised learning, \cite{tsai2020learning} introduce consistency regularization, and propose LLP-VAT, while the KL divergence on the proportions is intact in the loss.

However, KL divergence based methods hardly delivery classifiers abiding the proportions on training data.
When performing alternate update, this inexact labeling exaggerates the label noise, especially with entropic regularization.
In contrast, we address this problem with a constrained optimization to exactly follow the proportions.
Furthermore, the diversity in unlabeled data demonstrates useful behavior for classification \cite{lu2018minimal}.
Based on the great capacity of deep models, unsupervised representation learning is achievable and appealing as a promising step towards discriminative tasks.
For example, aligning the clustering with semantic classes can successfully improve image classification and segmentation \cite{ji2019invariant}.
\cite{asano2019self} apply (entropic constrained) optimal transport (OT) to learn representation with a self-labeling mechanism. Following their seminal framework, in this paper, we propose an OT based LLP algorithm, and further develop detailed techniques.

\section{Preliminaries}
In order to clearly describe how to leverage OT for LLP, we first introduce several preliminaries for both OT and LLP. More details are in Appendix due to the space limitation.

\subsection{Optimal transport via Kantorovich relaxation}
Define two discrete measures on two arbitrary sets $\mathcal{X}$ and $\mathcal{Y}$, respectively:
$\alpha\!=\!\sum_{i=1}^n a_i\delta_{x_i}, \beta\!=\!\sum_{j=1}^m b_j\delta_{y_j}$,
where $\boldsymbol{a}\!=\!(a_i)_{i=1}^n\!\in\!\Sigma^n$, $\boldsymbol{b}\!=\!(b_j)_{j=1}^m\!\in\!\Sigma^m$, and $\Sigma^s$ is probability simplex, i.e., $\Sigma^s\!=\!\{\boldsymbol{a}\!\in\!\mathds{R}_+^s\big|\sum_{i=1}^s a_i\!=\!1\}, ,\forall s\!\in\!\mathds{Z}_+$.
To achieve mass splitting in \emph{Kantorovich relaxation}, instead of permutation $\sigma$ or surjective map $T$, we use \emph{coupling matrix} $\mathbf{P}\!\in\!\mathds{R}_+^{n\times m}$ with $P_{ij}$ as the mass flowing from $x_i$ to $y_j$.
Admissible couplings give a simpler characterization than maps in Monge problem: Suppose $\boldsymbol{a}\in\Sigma^n, \boldsymbol{b}\in\Sigma^m$, we denote $U(\boldsymbol{a},\boldsymbol{b})\!=\!\{\mathbf{P}\in\mathds{R}_+^{n\times m}\mid\mathbf{P}\mathds{1}_m\!=\!\boldsymbol{a},\ \mathbf{P}^\intercal\mathds{1}_n\!=\!\boldsymbol{b}\}$,
where $\mathds{1}_{n}$ represents $n$-dimensional all ones column vector.
Hence, $U(\boldsymbol{a},\boldsymbol{b})$ stands for the set of all Kantorovich mappings.
Similar to Monge problem, let $\langle\cdot,\cdot\rangle$ be the Frobenius dot-product, given $\boldsymbol{a}, \boldsymbol{b}\in\Sigma^n$ and a cost matrix $\mathbf{C}$, the Kantorovich's \emph{optimal transport} (OT) problem is a linear programming as
\begin{equation}\label{ot}
\vspace{-3pt}
\mathbf{L}_{\mathbf{C}}(\boldsymbol{a},\boldsymbol{b})=\min_{\mathbf{P}\in U(\boldsymbol{a},\boldsymbol{b})}\langle\mathbf{P},\mathbf{C}\rangle=
\sum_{i,j}P_{ij}C_{ij}.
\vspace{-3pt}
\end{equation}

\subsection{Entropic regularization of optimal transport}
The solution of the original OT problem \eqref{ot} is non-unique and tends to be sparse, i.e., arriving at certain vertex of the polytope $U(\boldsymbol{a},\boldsymbol{b})$.
To form a more ``blurred'' prediction, \cite{cuturi2013sinkhorn} propose the entropic regularization for OT problem. To be specific, the discrete entropy of a coupling matrix $\mathbf{P}$ is well-known as $\mathbf{H}(\mathbf{P})=-\sum_{i,j}P_{ij}\log(P_{ij})$.

\begin{rem}[\cite{peyre2019computational}]
\label{rem1}
The entropic function $\mathbf{H}(\cdot)$ is 1-strongly concave, due to the negative definite Hessian matrix: $\partial^2\mathbf{H}(\mathbf{P})=-\mathbf{diag}(1./\mathbf{P})$ and $P_{ij}\leqslant 1, \forall i,j$.
\end{rem}
Note that the rank-1 matrix $\boldsymbol{a}\boldsymbol{b}^\intercal$ is an admissible coupling, and $\mathbf{H}(\mathbf{P})\!\leqslant\!\mathbf{H}(\boldsymbol{a})\!+\!\mathbf{H}(\boldsymbol{b})\!=\!\mathbf{H}(\boldsymbol{a}\boldsymbol{b}^\intercal)$.
The \emph{entropic regularization} of OT adds $-\mathbf{H}(\cdot)$ to original OT \eqref{ot} as:
\begin{equation}\label{er}
\mathbf{L}^\varepsilon_\mathbf{C}(\boldsymbol{a},\boldsymbol{b})=
\min_{\mathbf{P}\in U(\boldsymbol{a},\boldsymbol{b})}\langle\mathbf{P},\mathbf{C}\rangle
-\varepsilon\mathbf{H}(\mathbf{P}),
\vspace{-3pt}
\end{equation}
which is a constrained minimization problem of an $\varepsilon$-convex function (Remark \ref{rem1}), thus has a unique minimum.
More importantly, the entropic constraint guarantees a computationally efficient process to find the solution, as a consequence of restricting the search for low cost
joint probabilities within sufficient smooth tables \cite{cuturi2013sinkhorn}. 
Apparently, if $\varepsilon\!\rightarrow\!+\infty$, the solution of entropic regularized OT \eqref{er} is $\boldsymbol{a}\boldsymbol{b}^\intercal$.
In addition,
the solution of \eqref{er} converges to 
that of the original OT \eqref{ot} as $\varepsilon\rightarrow0$.

\subsection{Learning from label proportions}

In LLP problem, because the label proportions are available, we can restrict the instance-level self-labeling procedure with these proportional information using an OT framework. Before further discussion, we fist give the formal formulation for LLP by directly considering a multi-class problem with $K$ classes in this paper.
With no prior knowledge, we further suppose that the training data consist of $N$ \emph{randomly generated} disjoint bags. Consequently, the training data can be expressed as $\mathcal{D}=\{(\mathcal{B}_i,\boldsymbol{p}_i)\}_{i=1}^m$,
where $\mathcal{B}_i = \{\mathbf{x}_{i,j}\}_{j=1}^{n_i}$ denotes the instances in the $i^{th}$ bag, and $\mathcal{B}_i \cap \mathcal{B}_j = \varnothing, \forall i \neq j$.
The $\boldsymbol{p}_i\in[0,1]^K$ and $n_i$ are the known ground-truth label proportions and the bag size of the $i^{th}$ bag, respectively.

\section{Approach}

\subsection{Linking OT to LLP (PLOT)}
\cite{asano2019self} introduce a self-labeling framework to leverage equal clustering to learn discriminative representation on unsupervised data and achieve classification.
In LLP problem, although the class distribution is not uniform within each bag, we can easily modify
the admissible couplings in \eqref{ot}
to fit in the proportional information.
Consequently, with $p_i^y$ as the proportion of class $y$ in bag $i$, we have
\begin{equation}\label{bce}
\begin{split}
\min_{q}\ & BCE(p,q)\! =\!-\!\sum_{i=1}^m\!\sum_{j=1}^{n_i}\!\sum_{y=1}^K\! \frac{q(y|\mathbf{x}_{i,j})}{n_i}\!\log p_\phi(y|\mathbf{x}_{i,j})\\
s.t. & \sum_{j=1}^{n_i} q(y|\mathbf{x}_{i,j})= p_i^y\cdot n_i,\ q(y|\cdot)\!\in\![0,1],\\
&\forall y\!\in\!\llbracket K\rrbracket\!=\!\{1,2,\!\cdots\!,K\}, \forall i\!\in\!\llbracket m\rrbracket\!=\!\{1,2,\!\cdots\!,m\}.
\end{split}
\end{equation}
Note that \eqref{bce} is a constrained optimization and the labels should strictly comply with the proportions in each bag.
Nevertheless, \eqref{bce} will be combinatorial if imposing one-hot labeling in $q$ , thus is very difficult to optimize. Fortunately, as we will point out, \eqref{bce} is a typical OT problem, thus can be solved relatively efficiently to arrive at a sparse (one-hot) solution.

For better explanation, we rewrite \eqref{bce} in a matrix fashion.
Formally, let $\mathbf{Q}^i\!=\!(Q^i_{jk})\!\in\!\mathds{R}_+^{K\!\times\!n_i}, Q^i_{jk}\!=\!q(k|\mathbf{x}_{i,j})/n_i$, and $\mathbf{P}^i\!=\!(P^i_{jk})\!\in\!\mathds{R}_+^{K\!\times\!n_i}, P^i_{jk}\!=\!p_\phi(k|\mathbf{x}_{i,j})/n_i$. 
In addition, denote $\mathbf{Q}\!=\!\text{diag}\{\mathbf{Q}^i\}_{i=1}^m, \mathbf{P}\!=\!\text{diag}\{\mathbf{P}^i\}_{i=1}^m$, and $\boldsymbol{p}\!=\!(\boldsymbol{p}_1^\intercal,\boldsymbol{p}_2^\intercal,\cdots,\boldsymbol{p}_m^\intercal)^\intercal, \boldsymbol{b}\!=\!(\mathds{1}_{n_1}^\intercal/{n_1}, \mathds{1}_{n_2}^\intercal/{n_2}, \cdots, \mathds{1}_{n_m}^\intercal/{n_m})^\intercal$.
Define $U(\boldsymbol{p},\boldsymbol{b})\!=\!\big\{\mathbf{Q}\!\in\!\mathds{R}_+^{mK\!\times\!N}\mid\mathbf{Q}\mathds{1}_N\!=\!\boldsymbol{p},\ \mathbf{Q}^\intercal\mathds{1}_{mK}\!=\!\boldsymbol{b}\big\}$.
Accordingly, we have an equivalent OT problem for \eqref{bce} as:
\vspace{-5pt}
\begin{equation}\label{cotllp}
\min_{\mathbf{Q}\in U(\boldsymbol{p},\boldsymbol{b})}\langle\mathbf{Q}, -\log\mathbf{P}\rangle=BCE(p,q)+\log\prod_{i=1}^m n_i.
\vspace{-5pt}
\end{equation}
On the other hand, with $\lambda\!\rightarrow\!+\infty$, we can instead solve the entropic regularized OT problem to accelerate the process of convergence, as well as attaining unique non-sparse solution.
\vspace{-12pt}
\begin{equation}\label{erot}
\mathbf{L}^{1/\lambda}_\mathbf{-\log\mathbf{P}} (\boldsymbol{p},\boldsymbol{b})=\min_{\mathbf{Q}\in U(\boldsymbol{p},\boldsymbol{b})}\langle\mathbf{Q},-\log\mathbf{P}\rangle
-\frac{1}{\lambda}\mathbf{H}(\mathbf{Q}).
\end{equation}

\subsection{Alternating optimization}
In the proposed learning framework, the network parameters $\phi=(\varphi,\theta)$ and self-labels $\mathbf{Q}$ are alternately updated.
Now, we further describe the details as follows.
\paragraph{Training the network with fixed $\mathbf{Q}$.} Because the cross-entropy is differentiable w.r.t. the network parameters $\phi=(\varphi,\theta)$, we can directly conduct common optimizer, e.g., Adam, on the objective in \eqref{bce} by fixing $\mathbf{Q}$.
\paragraph{Updating the labels $\mathbf{Q}$ with fixed $\phi$.} When the model is fixed, the label assignment matrix $\mathbf{Q}$ are obtained by OT or entropic regularized OT.
When performing original OT, the solution $\mathbf{Q}^*$ is with 0/1 binary elements.
However, entropic regularized OT produces $\mathbf{Q}^*$ with all elements in [0,1].
In practice, we employ two strategies for label update: hard labeling and soft labeling.
In hard labeling, we update $\mathbf{Q}$ as:
\begin{equation}\label{hard-label}
Q^i_{js}\!=\! 
\left\{
\begin{aligned}
1, & \quad \text{if} \ \ s\!=\!\argmax_k \  
q(k|\mathbf{x}_{i,j})  \\
0, & \quad \text{otherwise}
\end{aligned}
\right.,\ i=1,2,\!\cdots\!,m.
\end{equation}
In soft labeling, we directly use the labels obtained by entropic regularized OT. In the experimental part, we provide the performance comparison on hard and soft labeling.

\subsection{Two-Stage training for LLP}

In practice, we consider to perform the clustering in every single bag, with the proportions as the constraint for instances number in each cluster. In detail, we conduct the constrained OT problem \eqref{cotllp} w.r.t. $\mathbf{Q}^i$ and $\mathbf{P}^i$, with minor revision on $U(\boldsymbol{p},\boldsymbol{b})$, i.e., $U(\boldsymbol{p}_i,\boldsymbol{b}_i)$, where $\boldsymbol{b}_i\!=\!\mathds{1}_{n_i}/{n_i}$.
On the other hand, we can instead solve the entropic regularized OT problem \eqref{erot} with the same revision as \eqref{erotllp} to accelerate the training, as well as obtain non-sparse solution to perform soft labeling:
\vspace{-5pt}
\begin{equation}\label{erotllp}
\mathbf{L}^{1/\lambda}_\mathbf{-\log\mathbf{P}} (\boldsymbol{p}_i,\boldsymbol{b}_i)\!=\!\min_{\mathbf{Q}^i\in U(\boldsymbol{p}_i,\boldsymbol{b}_i)}\langle\mathbf{Q}^i,-\log\mathbf{P}\rangle\!-\!\frac{1}{\lambda}\mathbf{H}(\mathbf{Q}^i).
\end{equation}

Our two-stage training process uses KL-divergence based LLP training as the first stage, and supervised learning with pseudo-labels generated by OT as the second stage.
In order to reduce the memorization of corrupt pseudo-labels, we incorporate symmetric cross-entropy (SCE) \cite{wang2019symmetric} in second stage.
Let $L_{CE}$ be cross-entropy $H(\boldsymbol{p},\boldsymbol{q})$, we denote $L_{RCE}$ as \emph{reverse} cross-entropy $H(\boldsymbol{q},\boldsymbol{p})$ defined as:
\vspace{-5pt}
\begin{equation}\label{rce}
\begin{split}
L_{RCE} = H(\boldsymbol{q},\boldsymbol{p})= - \sum_{k=1}^{K} p(k|x) \log q(k|x).
\vspace{-5pt}
\end{split}
\vspace{-5pt}
\end{equation}
Accordingly, the SCE is defined as
\begin{equation}\label{sce}
\begin{split}
SCE = L_{CE} + L_{RCE} = H(\boldsymbol{p},\boldsymbol{q}) + H(\boldsymbol{q},\boldsymbol{p}).
\end{split} 
\end{equation}

To further improve the performance, we apply mixup \cite{zhang2018mixup} to any two pairs $(\boldsymbol{x}_i,\boldsymbol{y}_i)$ and $(\boldsymbol{x}_j,\boldsymbol{y}_j)$, i.e.,  $\boldsymbol{x}\!=\! \lambda \boldsymbol{x}_i\!+\!(1\!-\!\lambda)\boldsymbol{x}_j$ and $\boldsymbol{y}\!=\!\lambda\boldsymbol{y}_i\!+\!(1\!-\!\lambda)\boldsymbol{y}_j$,
to obtain a new data point $(\boldsymbol{x},\boldsymbol{y})$.
The detailed training process of the second stage is shown in Algorithm \ref{alg:Framwork}. 

\section{Numerical experiments}
In order to demonstrate OL-LLP is model-agnostic, in this section, we conduct quantitative and qualitative studies with extensive experiments, to show the improvement of former LLP methods, when using PLOT as the second training stage.
We use two benchmark datasets: CIFAR-10 and CIFAR-100.
The comparisons are performed on three recently proposed algorithms DLLP \cite{ardehaly2017co}, LLP-VAT \cite{tsai2020learning}, and LLP-GAN \cite{liu2019learning}.

\begin{algorithm}
\caption{Second Stage Pseudo Labeling with OT (PLOT)}
\label{alg:Framwork}
\begin{algorithmic}[1]
\REQUIRE ~~\\
The LLP training data  $\mathcal{D}\!=\!\{(\mathbf{B}_i,\boldsymbol{p}_i)\}_{i=1}^m$;\\
The initialization of the network parameters: $\phi=(\varphi,\theta)$;\\
The block diagonal matrix $\mathbf{Q}=\text{diag}\{\mathbf{Q}^i\}_{i=1}^m$ as the pseudo-labels obtained in the first-stage.
\ENSURE ~~\\
\FOR{each epoch over $\mathcal{D}$ and the pseudo-labels $\mathbf{Q}$ }
\STATE Sample one mini-batch of training bags $B_D$ from $\mathcal{D}$ and the corresponding pseudo-labels from $\mathbf{Q}$ \\
\FOR{each mini-batch}
\STATE Train network based on SCE with the pseudo-labels to update the parameters  $\phi=(\varphi,\theta)$. 
\ENDFOR
\STATE Update the pseudo labeling matrix $\mathbf{Q}=\text{diag}\{\mathbf{Q}^i\}_{i=1}^m$ through the OT or entropic regularized OT.
\ENDFOR
\RETURN The appropriate values of $\phi=(\varphi,\theta)$.
\end{algorithmic}
\end{algorithm} 


\subsection{Synthetic datasets study}

We first evaluate our approach with a toy example.
In detail, we choose the well-known ``two-moon'' dataset to obtain the performance of DLLP and LLP-GAN, as well as their combination with PLOT as the second training stage.
A 3-hidden-layer neural network with ReLU activation is employed.
We conduct the experiment with the same bag setting, where 40 bags are generated with each containing 50 points.

\begin{figure}
\centering
\subfigure[DLLP]{
\centering
\includegraphics[width=0.24\textwidth]{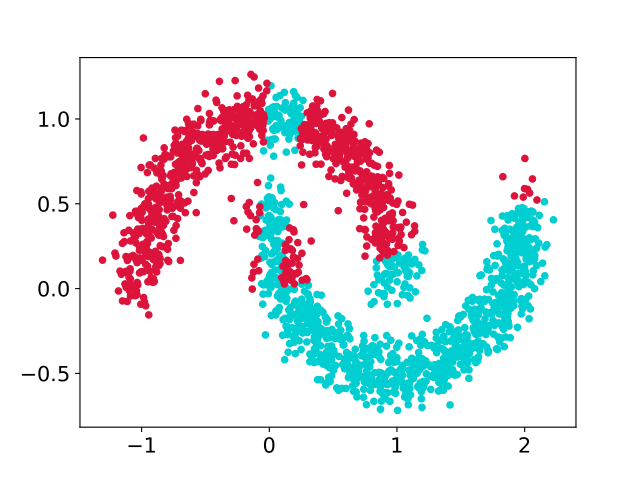}
\label{fig1_hyper}
}%
\subfigure[DLLP + PLOT]{
\centering
\includegraphics[width=0.24\textwidth]{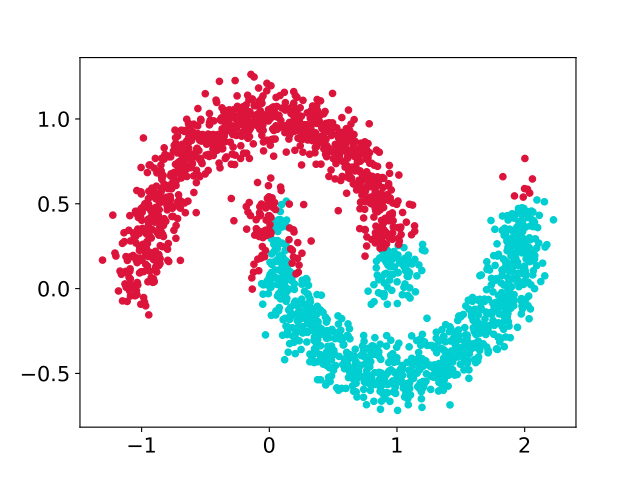}
\label{fig2_hyper}
}%
\\
\vspace{-12pt}
\centering
\subfigure[LLP-GAN]{
\centering
\includegraphics[width=0.24\textwidth]{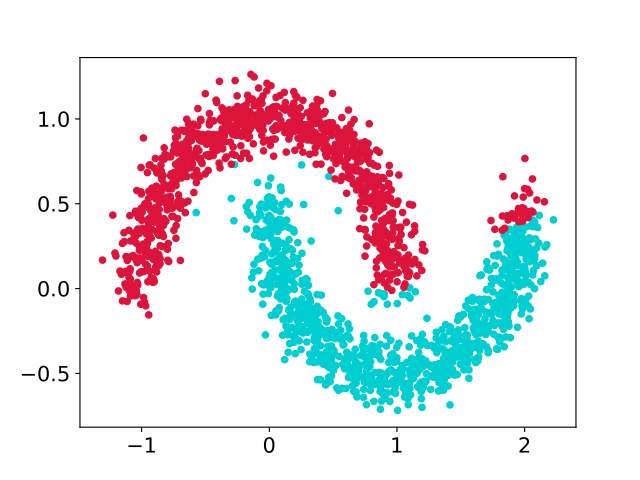}
\label{fig3_hyper}
}%
\subfigure[LLP-GAN + PLOT]{
\centering
\includegraphics[width=0.24\textwidth]{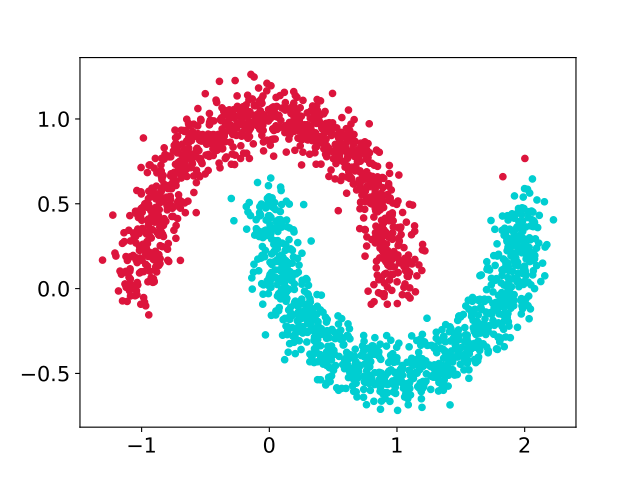}
\label{fig4_hyper}
}%
\vspace{-5pt}
\caption{Comparison of DLLP, LLP-GAN and their combinations with PLOT on the ``two moons'' dataset. (Best viewed in color)}
\label{result_synthetic}
\vspace{-5pt}
\end{figure}

The results are visualized in Figure \ref{result_synthetic}, with different classes in red and green, respectively. For DLLP in Figure \ref{fig1_hyper}, blobs of data is mis-classified for both categories, while the portion of errors reduces when incorporating PLOT to DLLP in Figure \ref{fig2_hyper}. When it comes to stronger baseline LLP-GAN in Figure \ref{fig3_hyper}, only the tails of both categories are mis-classified.
However, only visible two green points are mis-classified to red at the tail in Figure \ref{fig4_hyper}, whose results are obtained by conducting OP-LLP training as the second stage to LLP-GAN.
It effectively demonstrates that our second stage can further boost the performance of other LLP solvers.
In practice, solving OT can be regarded as a complementary refinement for the solution of minimizing the KL divergence.

\subsection{Experimental setting}

\paragraph{Label proportions generation.} 
As there is no off-the-shelf LLP datasets, we first generate LLP datasets with the label proportions using four supervised benchmark datasets.
Following the setting from \cite{liu2019learning}, we construct four kinds of bags, with bag sizes of 16, 32, 64, and 128, respectively. In order to avoid the influence of different label distributions, the bag setting is fixed across different algorithms.

\paragraph{Training setting.}
We choose a conv-based 13-layer max-pooling network as the backbone for all methods.
The architecture details is given in Appendix \ref{networkarchitecture}.
Meanwhile, Adam optimizer is used with $\beta_1\!=\!0.5$ and $\beta_2\!=\!0.999$.
The initial learning rate is $1e\!-\!4$, divided by $2$ every $100$ epochs.
Random horizontal flip and random crop with padding to the original images is employed as data augmentation.
The hyper-parameters for SCE is the same as \cite{wang2019symmetric}.

\subsection{Comparison with the state-of-the-arts}

In this section, we provide the overall accuracy comparison between our approach and three SOTA methods: DLLP, LLP-GAN, and LLP-VAT, on CIFAR-10 and CIFAR-100.
Specifically, our method divides the training into two phases, where the first phase is to train the three KL-divergence based models as the \emph{teacher model}, and the second stage is to update the \emph{student model} based on the pseudo-labels obtained in the first step, followed by solving the OT with the proportional information in an alternate manner, described in Algorithm \ref{alg:Framwork}.
In practice, note that our approach is model-agnostic, which means the first stage can be any deep learning based model, and our model can boost their performance by attaining local solutions with significantly better performance.


Tabel \ref{tab_cifar10} provides the final results on CIFAR-10, where for fair comparison, we report the results without mixup \cite{zhang2018mixup}.
For the first glance, because bigger bag size contains less supervised information when the total number of training data is fixed, the accuracies of all the methods decrease along with the increase of bag size.
Nevertheless, the performance of all methods with all bag sizes can efficiently improve when combining with the proposed second-stage training.
Note that the improvement is more obvious with the bag size of 64 and 128, which is more challenging compared with smaller bag sizes.
For example, the accuracy of  LLP-GAN + PLOT with bag size of 128 reaches 79.08, which is even superior to the previous SOTA results with bag size 64.
Furthermore, on CIFAR-10, this combination can approximately achieve a similar performance to the fully supervised scenario with small bag size, e.g., 16.
In general, we can further draw a conclusion: Algorithm with the best performance in the first stage consistently results in the best performance when introducing PLOT as the second stage.

\begin{table*}[htbp]
\centering
\begin{tabular}{ccccc}
  \specialrule{0.15em}{0pt}{2pt}
  \multirow{2}{*}{Model}  & \multicolumn{4}{c}{Bag Size} \\
\cmidrule{2-5}
  ~ &  16 & 32 & 64 & 128  \\
  \specialrule{0.1em}{0.5pt}{2pt}
   DLLP \cite{ardehaly2017co} & 87.69 (0.54) & 82.88 (0.47) & 71.11 (0.56) & 46.68 (0.75) \\
  DLLP + PLOT (\textbf{ours}) & \textbf{90.01} (0.23) &86.87 (0.24) & 79.69 (0.44) & 55.21 (0.56) \\
  \specialrule{0.1em}{2pt}{2pt}
   LLP-VAT \cite{tsai2020learning}  &88.36 (0.29) &83.01 (0.46) &70.53 (0.98) &51.25 (0.88) \\
   LLP-VAT + PLOT (\textbf{ours}) & 89.87 (0.31) & 87.89 (0.29) & 79.33 (0.39) & 63.28 (0.43) \\
  \specialrule{0.1em}{2pt}{2pt}
 LLP-GAN \cite{liu2019learning} & 86.97 (0.42) & 83.13 (0.43) & 77.21 (0.39) & 68.37 (1.21) \\
  LLP-GAN + PLOT (\textbf{ours}) & 89.15 (0.35) & \textbf{88.21} (0.23) & \textbf{84.14} (0.68) & \textbf{79.09} (0.44) \\
  \specialrule{0.15em}{1.5pt}{0pt}
\end{tabular}
\vspace{-5pt}
\caption{Test accuracy rates and standard deviations (\%) on CIFAR10 with different bag sizes. The results are obtained for 5 runs.}
\vspace{-5pt}
\label{tab_cifar10}
\end{table*}

\begin{table*}[htbp]
\centering
\begin{tabular}{ccccc}
  \specialrule{0.15em}{0pt}{2pt}
  \multirow{2}{*}{Model}  & \multicolumn{4}{c}{Bag Size} \\
  \cmidrule{2-5}
  ~ &  16 & 32 & 64 & 128  \\
  \specialrule{0.1em}{0.5pt}{2pt}
   DLLP \cite{ardehaly2017co} & 61.95 (0.54) & 55.22 (0.77) & 41.21 (1.36) & 15.17 (1.07) \\
  DLLP + PLOT (\textbf{ours}) & 65.20 (0.41) & 61.37 (0.74) & 49.55 (0.86) & 23.10 (0.29) \\
  \specialrule{0.1em}{2pt}{2pt}
   LLP-VAT \cite{tsai2020learning}  &65.21 (0.39) &58.65 (0.43) &45.53 (1.09) &21.25 (0.81) \\
   LLP-VAT + PLOT (\textbf{ours}) & 65.39 (0.28) & \textbf{62.01} (0.30) & 52.75 (0.51) & 28.23 (0.4) \\
  \specialrule{0.1em}{2pt}{2pt}
 LLP-GAN \cite{liu2019learning} & 61.66 (0.49) & 56.78 (0.55) & 50.29 (1.12) & 33.65 (1.25) \\
  LLP-GAN + PLOT (\textbf{ours}) & \textbf{65.41} (0.35) & 61.68 (0.48) & \textbf{55.66} (0.41) & \textbf{43.44} (0.71) \\
  \specialrule{0.15em}{1.5pt}{0pt}
\end{tabular}
\vspace{-5pt}
\caption{Test accuracy rates and standard deviations (\%) on CIFAR100 with different bag sizes. The results are obtained for 5 runs.}
\vspace{-5pt}
\label{tab_cifar100}
\end{table*}

When it comes to harder CIFAR-100 in Table \ref{tab_cifar100}, we have similar results: Our method can considerably boost other LLP methods to reach new state-of-the-art results.
In particular, our method reaches an accuracy rate of 43.44 with bag size 128, which is much higher than all the previous state-of-the-arts.
This indicates that incorporating the OT based pseudo-labels mechanism to learn label proportions is of higher efficiency.
In practice, the proportion of pseudo labeling can be intact to the real proportions through the optimal transport, thus improving the total accuracy of training data.

Note that the improvement for LLP-VAT is relatively limited compared to DLLP  and LLP-GAN when incorporating our post-hoc training, especially for bag sizes of 16 and 32.
A main reason is the key idea for noise reduction in network prediction is coincident in both methods, although different mechanisms are applied: LLP-VAT leverages the prediction consistency with adversarial samples as regularization, while our recipe is to obtain pseudo-labels with OT, then employs SCE and mixup as the post-hoc noise reduction techniques. 

\begin{figure}
\centering
\subfigure[CIFAR-10]{
\centering
\includegraphics[width=0.235\textwidth]{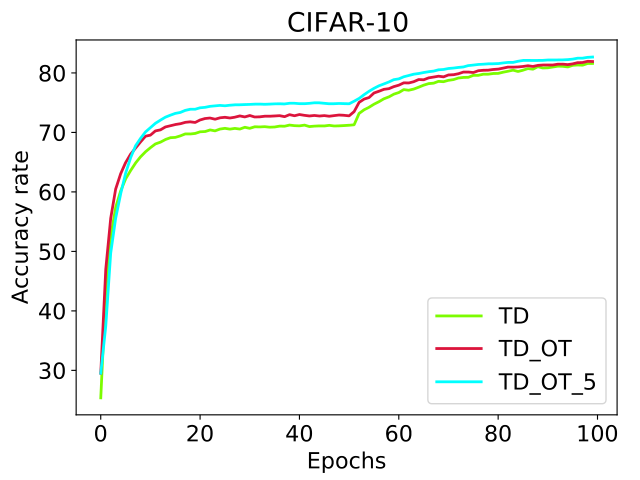}
}%
\subfigure[CIFAR-100]{
\centering
\includegraphics[width=0.235\textwidth]{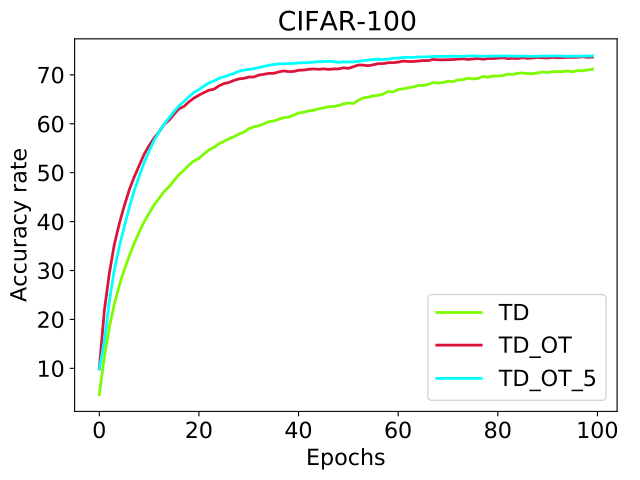}
}%
\vspace{-5pt}
\caption{Comparison of accuracy on CIFAR-10 and CIFAR-100 in the training process. (Best viewed in color)}
\label{result_training_acc}
\vspace{-5pt}
\end{figure}

The motivation behind our approach is the predicted labels of training data can be further refined by solving optimal transport, so as to fulfill better pseudo labeling.
To show this, we provide the accuracy curves for CIFAR-10 and CIFAR-100 with bag size 64 in Figure \ref{result_training_acc}, where the TD, TD$\_$OT, and TD$\_$OT$\_$5 denote the accuracies of the training data: without OT, with OT, and with the average of the last five pseudo-labels with OT (i.e., ensembles), respectively. 
It demonstrates that the accuracy of training data can significantly improve after involving optimal transport.
In particular, the advantage is more obvious for CIFAR-100, by a larger margin between TD and TD$\_$OT.
Meanwhile, the accuracy can further improve by using the average of the last five OT pseudo-labels 
as the final pseudo-label in every OT labeling phase.  

\subsection{Ablation study}
In this section, we carry out ablation study on key components introduced in PLOT.
All the experiments are conducted on the dataset CIFAR-10 with the bag size 64.
More specifically, our analysis mainly focuses on the difference between cross-entropy and symmetric cross-entropy.
Furthermore, we evaluate the effect of mixup augmentation to our model when updating network parameters with above losses.

The final results are shown in Table \ref{tab_ablation}.
The baseline is the performance of DLLP in the first stage.
The $L_{CE}$  and $L_{SCE}$ means that the model is trained without optimal transport, merely using the mean of instance-level predictions in the previous five epochs as pseudo-labels.
This strategy can also improve the performance slightly.
In particular, compared with $L_{CE}$, $L_{SCE}$ provides a $2\%$ higher performance improvement.
On the other hand, the accuracy can be further increased when employing the pseudo-labels generated by OT.
In particular, SCE is considerably effective, in contrast to CE, mainly due to its robustness to label noise.

Besides, mixup can further improve the performance significantly, which is mainly because of the following two reasons.
On one hand, it reduces the memorization of corrupt labels, similar to SCE.
On the other hand, it can impose more generalization on the network with the convex combination of a pair of examples, as well as their pseudo-labels.
\vspace{-5pt}
\begin{table}[htbp]
\centering
\begin{tabular}{cc}
  \specialrule{0.15em}{0pt}{1.5pt}
   Method  &  CIFAR-10 \\
  \specialrule{0.08em}{0.2pt}{1.2pt}
   baseline & 71.11 (0.36) \\
  \specialrule{0.05em}{0.2pt}{1.2pt}
   $L_{CE}$  &72.27 (0.39)  \\
  \specialrule{0.0em}{0.5pt}{0.5pt}
   $L_{SCE}$  &74.21 (0.24)  \\
  \specialrule{0.0em}{1pt}{1pt}
   $L_{CE}$ + OT    &78.57 (0.32)  \\
  \specialrule{0.0em}{0.5pt}{0.5pt}
  $L_{SCE}$ + OT   & 79.69 (0.44) \\
  \specialrule{0.0em}{0.5pt}{0.5pt}
  $L_{CE}$ + OT + MIXUP  &84.39 (0.22) \\
  \specialrule{0.15em}{0.5pt}{0pt}
\end{tabular}
\vspace{-5pt}
\caption{Ablation study on CIFAR-10 with bag size 64: $L_{CE}$ (cross-entropy), $L_{SCE}$ (symmetric cross-entropy); OT (update pseudo-labels with optimal transport), MIXUP (mixup data augmentation).}
\label{tab_ablation}
\end{table}


\subsection{Hard-label vs. Soft-label}

\begin{figure}
\centering
\subfigure[CIFAR-10]{
\centering
\includegraphics[width=0.23\textwidth]{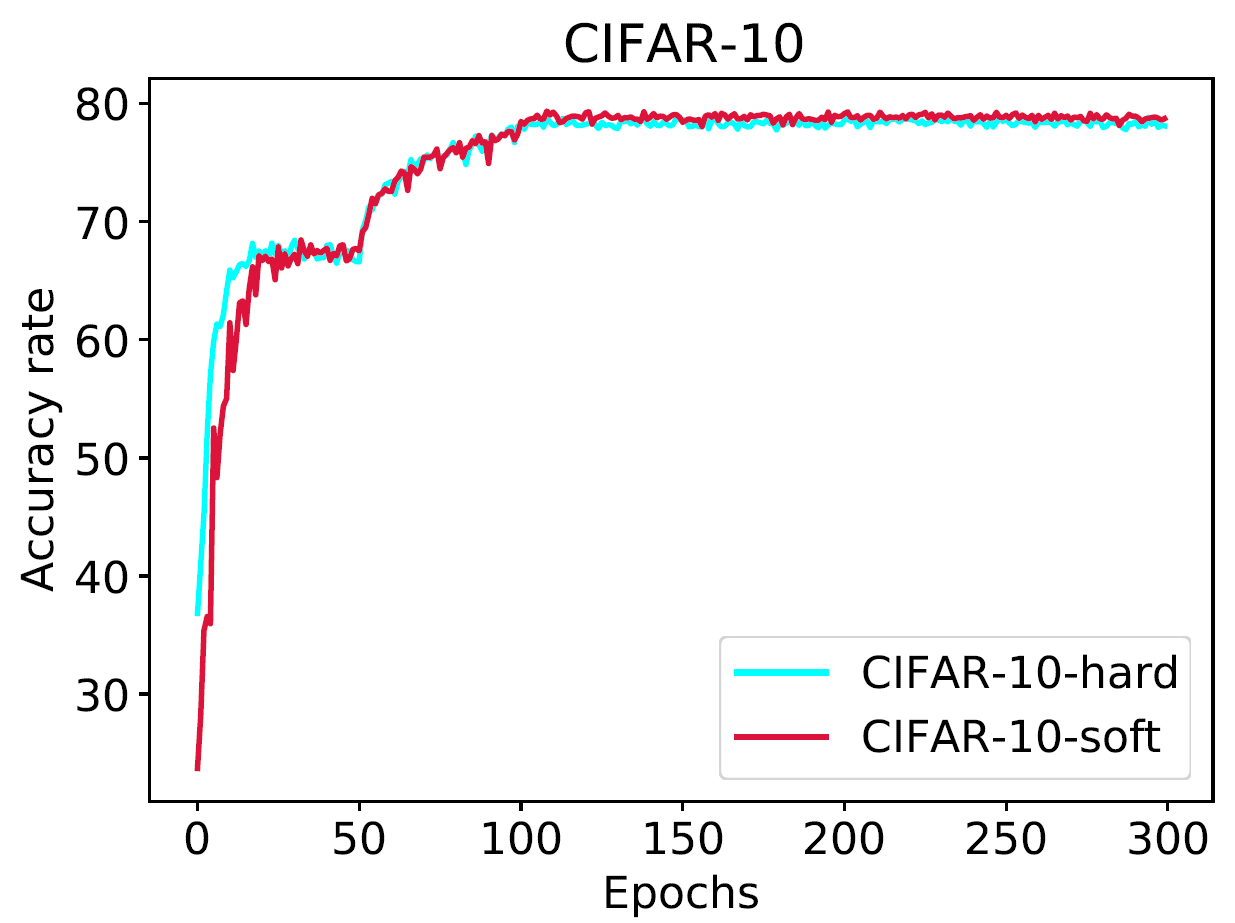}
}%
\subfigure[CIFAR-100]{
\centering
\includegraphics[width=0.23\textwidth]{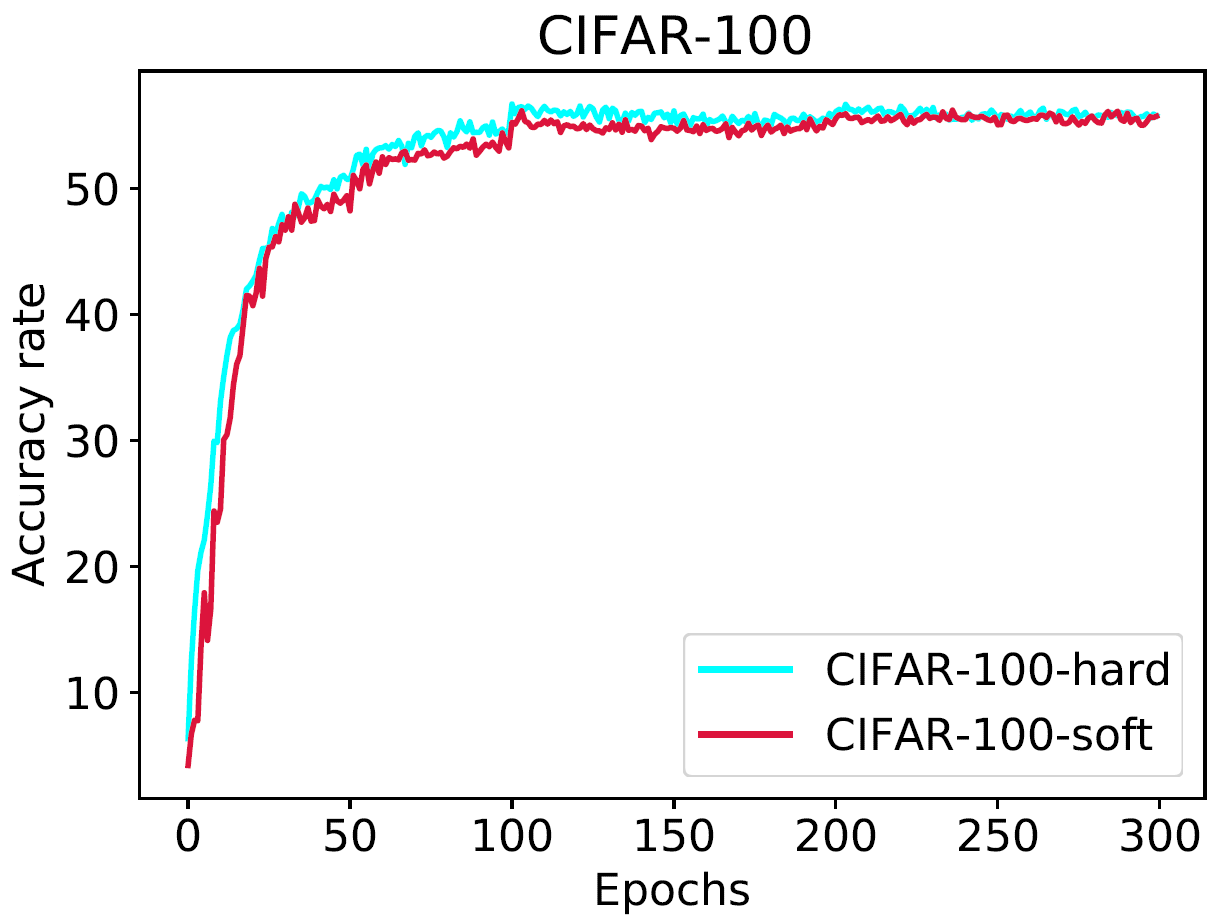}
}%
\vspace{-5pt}
\caption{Comparison of hard and soft pseudo labelings on two benchmarks. (Best viewed in color)}
\label{result_soft_hard}
\vspace{-5pt}
\end{figure}

In our model, we can employ two strategies to update the label: the hard labeling with the sparse original OT solution and the soft labeling with entropic regularized OT solution.
The detailed process is shown in \eqref{hard-label}.
In order to better demonstrate the difference between the hard-labels and soft-labels in our approach, we adopt CIFAR-10 and CIFAR-100 to conduct comparison on the performance with different bag sizes.
Specifically, the solution of the first stage is fixed for both hard and soft labeling for fair comparison.
The results are shown in Figure \ref{result_soft_hard}, where we provide the convergence curve of the second stage under different pseudo labelings.

From the result, we observe that it reach a comparable performance with hard and soft labelings for CIFAR-10 and CIFAR-100.
Intuitively, the soft-labels are more informative than the hard labels.
However, it does not necessarily lead to better performance due to the potential noisy labeling.
On the other hand, the solution of the original OT tends to be sparse, and with less noise.
We reckon this as the main reason of the comparable performance of two psuedo labeling strategies.
\vspace{-5pt}

\section{Discussion and Conclusion}\label{dc}
In this paper, we study the challenge in existing LLP methods, and point out the minimization on KL divergence between the prior and posterior class distributions is inadequate to comply with the label proportions exactly.
From this perspective, we propose a second stage for existing LLP solver with a framework to combine instance-level classification and pseudo labeling, and alternately optimize these two objectives.

Compared with former LLP solvers, especially \cite{dulac2019deep}, our main improvements are four-fold.
First, we introduce exact pseudo labeling to convert the unconstrained optimization based on KL divergence into a constrained one, so that the resulting labeling can strictly meet the proportions and avoid suboptimal solutions.
Second, although the pseudo-labels can be efficiently found by Sinkhorn’s algorithm for entropic regularized OT (or linear programming for OT), we instead recognize the supervised learning with the pseudo-labels as a noisy label problem, due to the insufficient proportional information.
Accordingly, we apply mix-up data augmentation and use SCE instead of CE as the loss function, which is differentiable and solved with a line search optimizer (e.g., Adam).
Third, we propose a two-stage LLP training by integrating PLOT as the second phase, and elaborately demonstrate that our framework can further improve the performance of DLLP, LLP-VAT and LLP-GAN with extensive experiments, thus is model-agnostic.
Fourth, instead of directly using the pseudo-labels, we empirically study the difference of hard and soft labeling strategies, and provide suggestions for the practical usage.

\newpage
\appendix
\section{Appendix}
In the appendix part, we firstly introduce some basic concepts for optimal transport, such as the Monge problem and its Kantorovich relaxation.
We also describe the Kantorovich mappings as a linear transformation (matrix), as well as the discussion on p-Wasserstein distance.
Besides, we introduce the  convergence of entropic regularized OT, which can be efficiently solved and has a close relation with the original OT.
Then, we link the supervised learning to unsupervised learning in a self-labeling and representation learning framework.
We point that all this content can be found in the references, and we reorganize them in the appendix to make our work self-contained.
We also provide the algorithm of Pseudo labeling with the Entropic regularized OT (PEOT) and the structure of our backbone.

\subsection{Monge problem}\label{monge}
First, define two discrete measures on two arbitrary sets $\mathcal{X}$ and $\mathcal{Y}$, respectively, i.e.,
$$\alpha=\sum_{i=1}^n a_i\delta_{x_i},\ \ \ \ \beta=\sum_{j=1}^m b_j\delta_{y_j},$$
where $\boldsymbol{a}\!=\!(a_i)_{i=1}^n\in\Sigma^n$ and $\boldsymbol{b}\!=\!(b_j)_{j=1}^m\in\Sigma^m$.
We continue to define a \emph{surjective} map $T\!:\!\mathcal{X}\!\rightarrow\!\mathcal{Y}$ that associates each $x_i$ to $y_j$, such that for any $j\!\in\!\llbracket m\rrbracket\!=\!\{1,2,\cdots,m\}$, $b_j\!=\!\sum_{Tx_i\!=\!y_j}\!a_i$.
Induced by mass transport, a transport of discrete probability measures is denoted in a compact \emph{push-forward} form $T_\sharp\alpha\!=\!\beta$. Furthermore, the \emph{Monge problem} is to seek an optimal $T$ to minimize the transportation cost w.r.t. a non-negative cost function $c(x,y)$ (satisfying the three properties of a legitimate distance metric) defined on $\mathcal{X}\!\times\!\mathcal{Y}$, i.e.,
\begin{equation}\label{mp}
\min_{T}\Big\{\sum_i c(x_i,Tx_i)\big|T_\sharp\alpha=\beta\Big\}.
\end{equation}
In particular, when $n=m$, then $T$ is a bijection from \emph{source} $\mathcal{X}$ to \emph{target} $\mathcal{Y}$, and can induce a permutation $\sigma\in\text{Perm}(n)$, such that $Tx_i=y_{\sigma(i)}$.

\begin{rem}\label{non-monge}
Note that the Monge problem may not even have a solution when the measures $\alpha$ and $\beta$ are not compatible, which is always the case when the target measure has more points than the source measure, i.e., $n<m$.

Besides, it is also the case when $n=m$ and there is at least one points in the target measure with the mass that does not match any point the source measure, i.e., $\exists\ b_j, s.t.\ b_j\neq a_i, \forall i$, and vice versa. We leave more complex situations for the readers due to the deviation from our main topic.
\end{rem}

Different from that mass transportation should be \emph{deterministic}, \emph{Kantorovich relaxation} instead considers a \emph{probabilistic} transport, which allows \emph{mass splitting} from a source toward several targets \cite{villani2008optimal}. Note that in this way, we may get rid of the challenge in the Monge problem in Remark \ref{non-monge}, because the source measure is regarded as the atom in the Monge problem while separable in Kantorovich relaxation w.r.t. the source measure.

\subsection{p-Wasserstein distance}
Suppose that $U(\boldsymbol{a},\boldsymbol{b})\!=\!\{\mathbf{P}\in\mathds{R}_+^{n\times m}\mid\mathbf{P}\mathds{1}_m\!=\!\boldsymbol{a},\ \mathbf{P}^\intercal\mathds{1}_n\!=\!\boldsymbol{b}\}$ is the set of all \emph{Kantorovich mappings},
where $\mathds{1}_{n}$ represents the column vector of all ones with dimension $n$. We have the following result.
\begin{rem}[\cite{peyre2019computational}]
The Kantorovich mapping is symmetric, i.e., $$\mathbf{P}\in U(\boldsymbol{a},\boldsymbol{b})\Leftrightarrow\mathbf{P}^\intercal\in U(\boldsymbol{b},\boldsymbol{a}).$$
\end{rem}
An important feature of OT \eqref{ot} is that it induces a distance between two probability measures in $\Sigma^n$, which are both discrete in this paper, as soon as the cost matrix satisfies the properties of a legitimate distance.
\begin{prop}
[\cite{villani2008optimal}]
\label{dis}
Let $\mathbf{D}\in\mathds{R}_+^{n\times n}$ be a distance on $\llbracket n\rrbracket\!=\!\{1,2,\cdots,n\}$ and $p\!\geqslant\!1$. We can define the $p$-Wasserstein distance on $\Sigma^n$ as
\begin{equation}
W_p(\boldsymbol{a},\boldsymbol{b})=\mathbf{L}_{\mathbf{D}^p} (\boldsymbol{a},\boldsymbol{b})^{1/p}, \forall \boldsymbol{a},\boldsymbol{b}\in\Sigma^n.
\end{equation}
\end{prop}
\begin{rem}
In particular, we have 
\begin{equation}
W_1(\boldsymbol{a},\boldsymbol{b})=\min\limits_{\mathbf{P}\in U(\boldsymbol{a},\boldsymbol{b})} \mathbb{E}_{(\boldsymbol{x},\boldsymbol{y})\sim\mathbf{P}} [\|\boldsymbol{x}-\boldsymbol{y}\|]
\end{equation}
as a legitimate distance between two distributions, which is used in Wasserstein GAN, as a weaker distance than the Jensen-Shannon (JS) Divergence in the original GAN and Kullback-Leibler (KL) divergence in the maximizing likelihood estimation (MLE).
\end{rem}
\subsection{Convergence of entropic regularized OT}\label{convergence}
The following results given by \cite{peyre2019computational} (Proposition 4.1), indicating the relationship between the original OT problem \eqref{ot} and the entropic regularized OT \eqref{er}.
\begin{prop}[Convergence of $\varepsilon$]
The unique solution $\mathbf{P}_\varepsilon$ converges to
the optimal solution with maximal entropy within the set of all optimal solutions of the OT problem \eqref{ot}, namely
\begin{equation}\label{2}
\mathbf{P}_\varepsilon\mathop{\longrightarrow}\limits^
{\varepsilon\rightarrow 0}
\mathop{\arg\min}\limits_{\mathbf{P}\in U(\boldsymbol{a},\boldsymbol{b})}
\{-\mathbf{H}(\mathbf{P})\mid\langle\mathbf{P},\mathbf{C}\rangle=
\mathbf{L}_\mathbf{C}(\boldsymbol{a},\boldsymbol{b})\},
\end{equation}
so that in particular
$$\mathbf{P}_\varepsilon\mathop{\longrightarrow}\limits^
{\varepsilon\rightarrow 0}\mathbf{L}_\mathbf{C}(\boldsymbol{a},\boldsymbol{b}).$$
On the other hand, we have
$$\mathbf{P}_\varepsilon\mathop{\longrightarrow}\limits^
{\varepsilon\rightarrow\infty}\boldsymbol{a}\otimes\boldsymbol{b}
=\boldsymbol{a}\boldsymbol{b}^\intercal.$$
\label{conv}
\end{prop}

\begin{rem}
A key insight is, as $\varepsilon$ increases, the optimal coupling of the problem \eqref{er} becomes less and less sparse, which in turn has the effect of both accelerating computational algorithms, and leading to faster statistical convergence.

Besides, we can decrease the magnitude of $\varepsilon$ to restore the original OT problem.
In fact, the entropic regularization term provides a trade-off between the computational efficiency and the sparsity of solution.
\end{rem}

\subsection{From supervised learning to unsupervised learning}\label{fromto}
Most of the content in this section is the reorganization of the seminal work in \cite{asano2019self}. We present the most relevant results here as an essential preliminary, in order to help the readers to clearly understand our PLOT framework.

In the standard supervised learning, after mapped by a deep neural network (DNN) $\Phi_{\theta}(\cdot)$, the high-level representation $\Phi_{\theta}(\mathbf{x})$ is followed by the fully connected multi-layer perceptron (MLP) $h_{\varphi}$ to yield the class-specific logits.
Based on these logits, DNN obtains the corresponding posterior class probabilities with a \emph{softmax} operation. Denote the multi-class classification head by $$g_{\varphi}:\ \mathds{R}^D\rightarrow\mathds{R}^K, \ \ g_{\varphi}=\mathrm{softmax}\circ h_{\varphi}.$$
In other words, the probabilistic output for an instance $\mathbf{x}$ can be summarized as
\begin{equation}
p_{\phi}(\boldsymbol{y}|\mathbf{x})=g_{\varphi}(\Phi_{\theta}(\mathbf{x})) =\mathrm{softmax}\circ h_{\varphi}(\Phi_{\theta}(\mathbf{x}))\in 2^K,
\label{equation_self}
\end{equation}
where $\phi\!=\!(\varphi,\theta)$ is the parameters of the model, which are the head and representation parameters of DNN, respectively, and learned with training dataset: $$\mathcal{D}\!=\!\{(\mathbf{x}_1,y_1),(\mathbf{x}_2,y_2),\!\cdots\!,(\mathbf{x}_N,y_N)| (\mathbf{x}_i,y_i)\!\in\!\mathcal{X}\times\llbracket K\rrbracket\},$$
by minimizing the average cross-entropy w.r.t. $(\varphi,\theta)$:
\begin{equation}
\begin{split}
L_{CE}(y,p|\mathcal{D},\theta,\varphi)&\!=\! -\frac{1}{N}\!\sum_{i=1}^N \!\sum_{k=1}^K \mathds{I}(y_i\!=\!k)\!\log p(k|\mathbf{x}_i,\theta,\varphi)\\
&=\!-\frac{1}{N}\!\sum_{i=1}^N\!\log p(y_i|\mathbf{x}_i,\theta,\varphi). \label{supervised-ce}
\end{split}
\end{equation}

On the other hand, the standard semi-supervised learning normally leverages a transductive scheme to incorporate the information of unlabeled data through self-labeling framework.
The key idea is, when the labels are unavailable, we resort to a self-labeling mechanism to assign the pseudo-labels automatically.
To be concrete, suppose $\{(\mathbf{x}_1,y_1),(\mathbf{x}_2,y_2),\cdots,(\mathbf{x}_l,y_l)\}$ is the labeled data, and $\{\mathbf{x}_{l+1},\mathbf{x}_{l+2},\cdots,\mathbf{x}_{l+u}\}$ is the unlabeled data, i.e., $$\mathcal{D}=\{(\mathbf{x}_i,y_i)\}_{i=1}^l\bigcup\{\mathbf{x}_j\}_{j=l+1}^{l+u}.$$
The representation part and head parameters $(\varphi,\theta)$ are learned by minimizing the average cross-entropy $\mathcal{L}$ on $\mathcal{D}$
\begin{equation}\label{sm}
\begin{split}
\mathbb{E}_{P(X,Y)}(\mathcal{L}(g(X),Y)&\backsimeq \hat{\mathbb{E}}(\mathcal{L}(g(\mathbf{x}),y))\\
&=\!-\frac{1}{l\!+\!u}\! \sum_{i=1}^{l+u}\!\sum_{k=1}^{K}\!\mathds{I}(y_i=k)\!\log p_{\phi}^k(\mathbf{x}_i).
\end{split}
\end{equation}
Although the formulation remains the same in \eqref{supervised-ce} and \eqref{sm}, different to \eqref{supervised-ce} in supervised learning, the transductive scheme in semi-supervised learning is to proceed joint optimization on \eqref{sm} w.r.t. $\phi=(\theta,\varphi)$ and the labeling $\{y_i\}_{i=l+1}^{l+u}$ of the unlabeled data simultaneously, according to the fully supervised information in labeled data.
In other words, we can alternately conduct optimization on the following two problems by fixing the other parameters:
\begin{equation}
\begin{split}
\min_{(\varphi,\theta)}\ &-\frac{1}{l+u} \sum_{i=1}^{l+u}\log p_{\phi}^{y_i}(\mathbf{x}_i)\\
&=-\frac{1}{l+u} \sum_{i=1}^{l+u}\log (\mathrm{softmax}\circ h_{\varphi}(\Phi_{\theta}(\mathbf{x}_i))_{y_i},
\end{split}
\end{equation}
and
\begin{equation}
\begin{split}
\min_{\boldsymbol{y}\in \llbracket K\rrbracket^u}&\ \ \ \  -\!\frac{1}{u}\!\sum_{i=l+1}^{l+u} \!\sum_{k=1}^{K}\!\mathbb{I}(y_i\!=\!k)\!\log p_{\phi}^k(\mathbf{x}_i)\\
&=\!-\frac{1}{u}\! \sum_{i=l+1}^{l+u}\!\sum_{k=1}^{K}\!\mathbb{I}(y_i\!=\!k)\!\log (\mathrm{softmax}\!\circ\! h_{\varphi}(\Phi_{\theta}(\mathbf{x}_i))_{k},
\end{split}
\end{equation}
where $\boldsymbol{y}=(y_{l+1},y_{l+2},\cdots,y_{l+u})^\intercal$ is the unknown labels for the unlabeled data.

The above formulation can be naturally extended to fully unsupervised learning scenario, with training data $\mathcal{D}\!=\!\{\mathbf{x}_1,\mathbf{x}_2,\!\cdots\!,\mathbf{x}_N\}$, by performing the clustering and representation learning simultaneously. 
However, the intact transfer of this mechanism without constraints on labeling will lead to degeneration to the final result: Assigning all the data points to a single (arbitrary) label. To avoid this issue, \cite{asano2019self} introduce the following constrained self-labeling framework to solve the clustering.

First, the \emph{soft} encoding of the labels is to utilize the probabilistic (non-mutually-exclusive) classifier to inference the posterior distribution (pseudo-labeling), based on the following cross-entropy loss:
\begin{equation}
CE(p,q)=-\frac{1}{N}\sum_{i=1}^N\sum_{y=1}^K q(y|\mathbf{x}_i)\log p(y|\mathbf{x}_i).
\end{equation}

To be concrete, for fixed $p$, the constrained minimization problem of $q$ according to $CE(p,q)$ is
\begin{equation}\label{cce}
\begin{split}
\min_{q}\ \ & CE(p,q)\\
s.t.\ \ & \sum_{i=1}^N q(y|\mathbf{x}_i)=\frac{N}{K}\\
& q(y|\cdot)\in\{0,1\}, \forall y\in\llbracket K\rrbracket.
\end{split}
\end{equation}

Accordingly, we obtain a combinatorial programming and thus may appear very difficult to optimize.

Fortunately, we can transfer \eqref{cce} to a standard optimal transport problem, which is a linear programming, or solve it quickly, thanks to the Sinkhorn's algorithm \cite{cuturi2013sinkhorn}.
This can be concluded as the \textbf{self-labeling} step, or the \textbf{pseudo labeling} step in our current work.

Formally, let $\mathbf{Q}=(Q_{ij})\in\mathds{R}_+^{K\times N}$ and $Q_{ij}=q_{i}(\mathbf{x}_j)/N$. Similarly, let $\mathbf{P}=(P_{ij})\in\mathds{R}_+^{K\times N}$ and $P_{ij}=p_{i}(\mathbf{x}_j)/N$.

In addition, denote $\boldsymbol{a}=\frac{1}{K}\mathds{1}_K, \boldsymbol{b}=\frac{1}{N}\mathds{1}_N$, and $U(\boldsymbol{a},\boldsymbol{b})\!=\!\big\{\mathbf{Q}\in\mathds{R}_+^{K\times N}\!\mid\!\mathbf{Q}\mathds{1}_N\!=\!\boldsymbol{a},\ \mathbf{Q}^\intercal\mathds{1}_K\!=\!\boldsymbol{b}\big\}$.
Then, we can give an equivalent problem for \eqref{cce} with the following OT problem:
\begin{equation}\label{cot}
\min_{\mathbf{Q}\in U(\boldsymbol{a},\boldsymbol{b})}\langle\mathbf{Q}, -\log\mathbf{P}\rangle=CE(p,q)+\log N.
\end{equation}

After updating $q$ with the problem \eqref{cce}, we can perform unconstrained optimization on $CE(p,q)$ w.r.t. the parameter in $p$, i.e., $\min_{\phi} CE(p_\phi,q)$.
This can be denoted as the \textbf{representation learning} step.
In the end, we can alternately conduct the above \emph{self-labeling} and \emph{representation learning} steps, to realize the unsupervised classification.

\subsection{The algorithm based on the entropic regularized OT}
Based on the Sinkhorn’s algorithm for the entropic regularized OT \cite{cuturi2013sinkhorn}, we describe the pseudo labeling with the entropic regularized OT algorithm, PEOT, in Algorithm \ref{alg:LLP-EROT}, as a realization of PLOT framework.
Accordingly, if applying the original OT without entropic regularization, we can change step 3 and step 4 with standard OT optimization, which will introduce more computation workload.
\begin{algorithm}[htbp]
    \caption{Second Stage Pseudo labeling with the Entropic regularized OT (PEOT)}
    \begin{algorithmic}[1]\label{alg:LLP-EROT}
    \REQUIRE ~~\\ 
    The LLP training data $\mathcal{D}\!=\!\{(\mathbf{B}_i,\boldsymbol{p}_i)\}_{i=1}^m$;\\
    $\lambda\!\in\! (0,+\infty)$, the threshold $\varepsilon>0$, and $\delta>\varepsilon$;\\
    Initialize $\mathbf{P}^\Delta\!=\!\text{diag}\{\mathbf{P}^i\}_{i=1}^m\!=\!\text{diag}\{\mathds{1}_{K\!\times\!n_i}/(n_iK)\}_{i=1}^m$;\\ $\boldsymbol{v}^{(0)}=\mathds{1}_{n_i}$, and $\boldsymbol{b}_i=\frac{1}{n_i}\mathds{1}_{n_i}$.
    \WHILE{$\delta>\varepsilon$}
    \FOR{each $i\in\llbracket m\rrbracket$}
    \STATE Solve the entropic regularized OT problem \eqref{erotllp} by iteratively update to obtain $\mathbf{Q}^i$ for the assignment of bag $i$, using $\boldsymbol{Q}_l^i=\text{diag}\{\boldsymbol{u}^{(l)}\}\mathbf{K}^{\lambda}\text{diag}\{\boldsymbol{v}^{(l)}\}$, with    \begin{equation}\nonumber
    \begin{split}
    \mathbf{K}^\lambda&=\exp\{\lambda\log\mathbf{P}^i\}\\
    \boldsymbol{u}^{(l)}&=\boldsymbol{p}_i./\mathbf{K}^\lambda \boldsymbol{v}^{(l)}\\ \boldsymbol{v}^{(l+1)}&=\boldsymbol{b}_i./(\mathbf{K}^\lambda)^\intercal \boldsymbol{u}^{(l)}
    \end{split}
    \end{equation}
    \STATE $\boldsymbol{Q}^i\!=\!\lim\limits_{l\rightarrow+\infty}\boldsymbol{Q}^i_l$. (The convergence is element-wise and proved in \cite{peyre2019computational}, Remark 4.8.)
    \ENDFOR
    \STATE Combine $\{\mathbf{Q}^i\}_{i=1}^m$ as the diagonal element to obtain the block diagonal matrix $\mathbf{Q}=\text{diag}\{\mathbf{Q}^i\}_{i=1}^m$.
    \STATE Fixing $\mathbf{Q}$, solve the following unconstrained programming \eqref{pu} w.r.t. the network parameters $\phi=(\varphi,\theta)$ on the whole training data:
    \begin{equation}\label{pu}
    \min_{\phi=(\varphi,\theta)}\!L_{CE}(p_\phi,q)\!=-\frac{1}{N}\!\sum_{i=1}^N\!\sum_{y=1}^K\!q(y|\mathbf{x}_i)\!\log\! p_{\phi}(y|\mathbf{x}_i).
    \end{equation}
    \STATE $\delta=\|\mathbf{P}-\mathbf{P}^{\Delta}\|_F$.
    \STATE $\mathbf{P}^{\Delta}=\mathbf{P}$.
    \ENDWHILE
    \ENSURE The final network parameters $\phi=(\varphi,\theta)$.
\end{algorithmic}
\end{algorithm}

\subsection{The network architecture using in the experiments}\label{networkarchitecture}

In Table \ref{baseline-cifar-10}, we deliver the convolution neural network architecture used in our experiments for all the methods on both CIFAR-10 and CIFAR-100 datasets.
More specifically, we use a 13-layer convolutional neural network as our backbone, where Leaky ReLU is chosen as the activation function.
\begin{table}[htbp]
\vspace{-5pt}
\centering
\begin{tabular}{c}
\specialrule{0.15em}{0pt}{2pt}
Input 32$\times$32 RGB image \\
\specialrule{0.05em}{2pt}{2pt}
3$\times$3 conv. 128 followed by LeakyReLU \\
\specialrule{0em}{2pt}{2pt}
3$\times$3 conv. 128 followed by LeakyReLU \\
\specialrule{0em}{2pt}{2pt}
3$\times$3 conv. 128 followed by LeakyReLU \\
\specialrule{0.05em}{2pt}{2pt}
2$\times$2 max-pooling with stride 2, dropout with $p=0.5$ \\ 
\specialrule{0.05em}{2pt}{2pt}
3$\times$3 conv. 256 followed by LeakyReLU \\
\specialrule{0em}{2pt}{2pt}
3$\times$3 conv. 256 followed by LeakyReLU \\
\specialrule{0em}{2pt}{2pt}
3$\times$3 conv. 256 followed by LeakyReLU  \\
\specialrule{0.05em}{2pt}{2pt}
2$\times$2 max-pooling with stride 2, dropout with $p=0.5$ \\ 
\specialrule{0.05em}{2pt}{2pt}
3$\times$3 conv. 512 followed by LeakyReLU  \\
\specialrule{0em}{2pt}{2pt}
1$\times$1 conv. 256 followed by LeakyReLU \\
\specialrule{0em}{2pt}{2pt}
1$\times$1 conv. 128 followed by LeakyReLU \\
\specialrule{0.05em}{2pt}{2pt}
Global mean pooling \\
\specialrule{0.05em}{2pt}{2pt}
Dense 10 \\
\specialrule{0.05em}{2pt}{2pt}
10-way Softmax \\
\specialrule{0.15em}{0pt}{2pt}
\end{tabular}
\caption{The backbone architecture for CIFAR-10 and CIFAR-100.}
\label{baseline-cifar-10}
\vspace{-5pt}
\end{table}

\begin{table}[htbp]
\caption{The baseline's architectures for MNIST, K-MNIST and F-MNIST.}
\label{baselines}
\centering
\begin{tabular}{c}
\hline
Input 28$\times$28 gray image   \\
\hline
Dense 28$\times$28 $\rightarrow$ 1000 followed by ReLU  \\
Dense 1000 $\rightarrow$ 500 followed by ReLU  \\
Dense 500  $\rightarrow$ 250 followed by ReLU \\
Dense 250  $\rightarrow$ 250 followed by ReLU \\
Dense 250  $\rightarrow$ 250 followed by ReLU  \\ Dense 250 $\rightarrow$ 10 \\
\hline
10-way Softmax\\
\hline
\end{tabular} 
\end{table}

\newpage

\bibliographystyle{named}
\bibliography{ijcai21}

\end{document}